\documentclass[11pt,a4paper]{article}
\usepackage[hyperref]{acl2021}
\usepackage{times}
\usepackage{latexsym}

\usepackage{graphicx} 
\usepackage{amsfonts}
\usepackage{amsmath}
\usepackage{booktabs}
\usepackage{algorithm}
\usepackage{amssymb}
\usepackage{booktabs}
\usepackage{multirow}
\usepackage{array,booktabs,arydshln}
\usepackage[switch]{lineno}  %
\usepackage{amsmath,algorithm,tabularx}
\usepackage[noend]{algpseudocode}
\makeatletter
\newcommand{\multiline}[1]{%
  \begin{tabularx}{\dimexpr\linewidth-\ALG@thistlm}[t]{@{}X@{}}
    #1
  \end{tabularx}
}

\usepackage{microtype}

\aclfinalcopy 
\def\aclpaperid{1044} 

\title{Enhancing Content Preservation in Text Style Transfer\\Using Reverse Attention and Conditional Layer Normalization}

\author{Dongkyu Lee \qquad Zhiliang Tian \qquad Lanqing Xue \qquad Nevin L. Zhang \\
  Department of Computer Science and Engineering,\\
  The Hong Kong University of Science and Technology\\
  \texttt{\string{dleear, ztianac, lxueaa, lzhang\string}@cse.ust.hk}
}

\date{}

\begin{document}
\maketitle
\begin{abstract}
Text style transfer aims to alter the style (e.g., sentiment) of a sentence while preserving its content. A common approach is to map a given sentence to content representation that is free of style, and the content representation is fed to a decoder with a target style. Previous methods in filtering style completely remove tokens with style at the token level, which incurs the loss of content information. In this paper, we propose to enhance content preservation by implicitly removing the style information of each token with reverse attention, and thereby retain the content. Furthermore, we fuse content information when building the target style representation, making it dynamic with respect to the content. Our method creates not only style-independent content representation, but also content-dependent style representation in transferring style. Empirical results show that our method outperforms the state-of-the-art baselines by a large margin in terms of content preservation. In addition, it is also competitive in terms of style transfer accuracy and fluency.

\end{abstract}
\section{Introduction}
Style transfer is a popular task in computer vision and natural language processing. It aims to convert an input with a certain style (e.g., sentiment, formality) into a different style while preserving the original content.

One mainstream approach is to separate style from content, and to generate a transferred sentence conditioned on the content information and a target style. Recently, several models \cite{li-etal-2018-delete,xu-etal-2018-unpaired,ijcai2019-732} have proposed removing style information at the token level by filtering out tokens with style information, which are identified using either attention-based methods \cite{DBLP:journals/corr/BahdanauCB14} or frequency-ratio based methods \cite{ijcai2019-732}. This line of work is built upon the assumption that style is \textit{localized} to certain tokens in a sentence, and a token has \textit{either} content or style information, but \textit{not both}. Thus by utilizing a style marking module, the models filter out the style tokens entirely when constructing a style-independent content representation of the input sentence.  
\begin{figure}[t]
  \centering
  \includegraphics[width=\columnwidth]{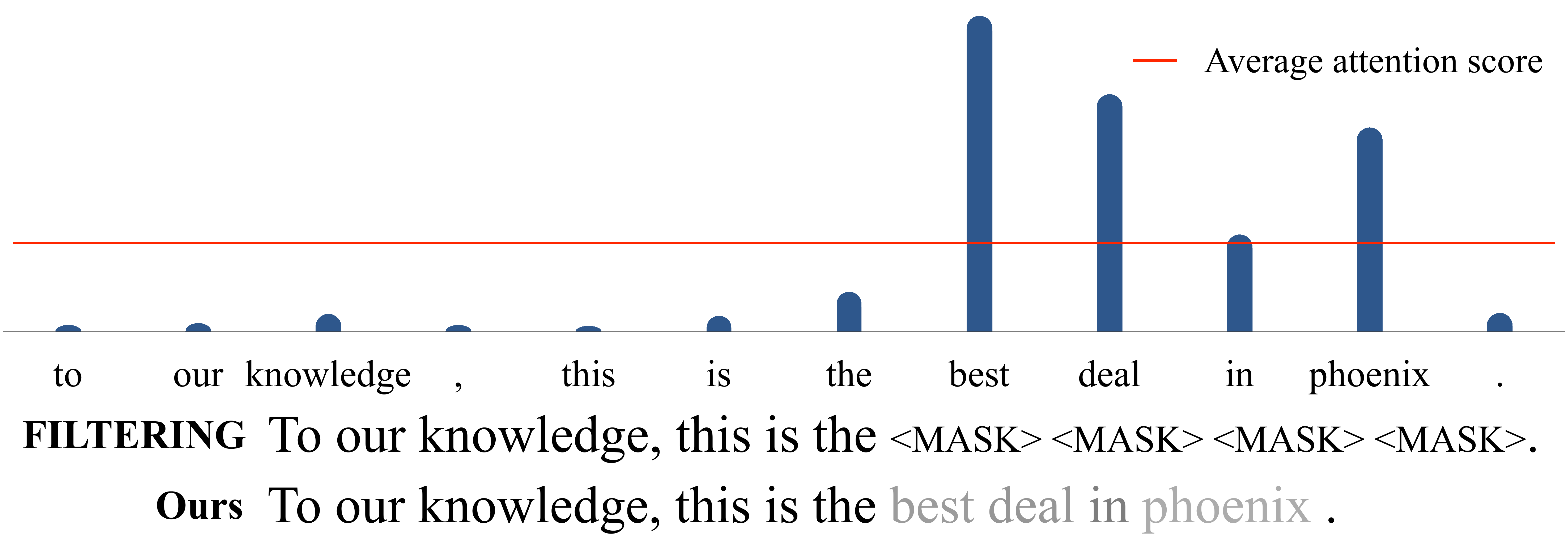}
  \caption{\small Illustration of difference between our method and filtering method in handling flat attention distribution. Each bar indicates attention score of the corresponding word.}
  \label{attentionExample}
  \end{figure}
The drawback with the filtering method is that one needs to manually set a threshold to decide whether a token is stylistic or content-related. Previous studies address this issue by using the average attention score as a threshold \cite{li-etal-2018-delete,xu-etal-2018-unpaired,ijcai2019-732}. A major shortcoming of this approach is the incapability of handling flat attention distribution. When the distribution is flat, in which similar attention scores are assigned to tokens, the style marking module would remove/mask out more tokens than necessary. This incurs information loss in content as depicted in Figure \ref{attentionExample}.

In this paper, we propose a novel method for text style transfer. A key idea is to exploit the fact that a token often posses both style and content information. For example, the word “delicious” is a token with strong style information, but it also implies the subject is food. Such words play a pivotal role in representing style (e.g., positive sentiment) as well as presenting a hint at the subject matter/content (e.g., food). The complete removal of such tokens leads to the loss of content information.

For the sake of enhancing content preservation, we propose a method to \textit{implicitly} remove style at the token level using \textit{reverse attention}. We utilize knowledge attained from attention networks \cite{DBLP:journals/corr/BahdanauCB14} to estimate style information of a token, and suppress such signal to take out style. Attention mechanism is known to attend to interdependent representations given a query. In style classification task, an attention score could be interpreted as to what extent a token has style attribute. If we can identify which tokens reveal stylistic property and to what extent, it is then possible to take the negation and to approximate the amount of content attribute within a token. In this paper, we call it reverse attention. We utilize such score to suppress the stylistic attribute of tokens, fully capturing content property.

This paper further enhances content preservation by fusing content information in creating target style representation. Despite of extensive efforts in creating content representation, the previous work has overlooked building content-dependent style representations. The common approach is to project the target style onto an embedding space, and share the style embedding among the same style as an input to the decoder. However, our work sheds light on building \textit{content-related style} by utilizing conditional layer normalization (CLN). This module of ours takes in content representations, and creates content-dependent style representation by shaping the content variable to fit in the distribution of target style. This way, our style representation varies according to the content of the input sequence even with the same target style. 

Our method is based on two techniques, Reverse Attention and Conditional Layer Normalization, thus we call it RACoLN. In empirical evaluation, RACoLN achieves the state-of-the-art performance in terms of content preservation, outperforming the previous state-of-the-art by a large margin, and shows competency in style transfer accuracy and fluency. The contributions are as follows:
\begin{itemize}
\item We introduce reverse attention as a way to suppress style information while preserving content information when building a content representation of an input.
\item Aside from building style-independent content representation, our approach utilizes conditional layer normalization to construct content-dependent style representation.
\item Our model achieves state-of-the-art performance in terms of content preservation, outperforming current state-of-the-art by more than 4 BLEU score on Yelp dataset, and shows competency in other metrics as well.
\end{itemize}

\begin{figure*}[t]
\centering
\includegraphics[width=0.9\textwidth]{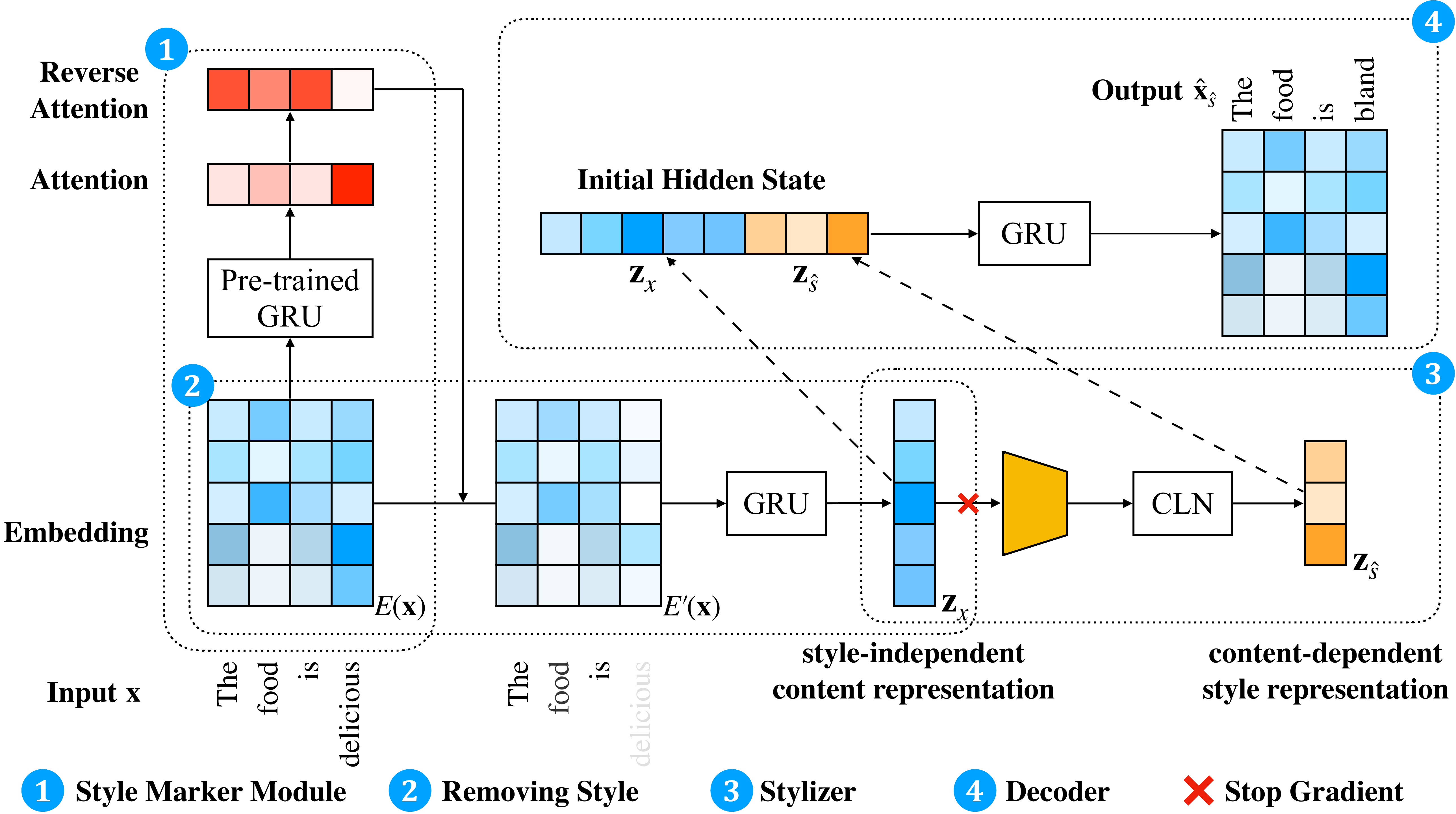}
\caption{\small Input $\mathbf{x}$ first passes style marker module for computing reverse attention. The reverse attention score is then applied to token embeddings, implicitly removing style. The content representation from the encoder is fed to stylizer, in which style representation is made from the content. The decoder generates transferred output by conditioning on the two representations.}
\label{modelStructure}
\end{figure*}
\section{Related Work}
In recent years, text style transfer in unsupervised learning environment has been studied and explored extensively. Text style transfer task views a sentence as being comprised of content and style. Thus, there have been attempts to disentangle the components \cite{NIPS2017_7259, li-etal-2018-delete,xu-etal-2018-unpaired,ijcai2019-732}. \newcite{NIPS2017_7259} map a sentence to a shared content space among styles to create style-independent content variable. Some studies view style as localized feature of sentences. \newcite{xu-etal-2018-unpaired} propose to identify style tokens with attention mechanism, and filter out such tokens. Frequency-based is proposed to enhance the filtering process \cite{ijcai2019-732}. This stream of work is similar to our work in that the objective is to take out style at the token level, but different since ours does not remove tokens completely. 

Instead of disentangling content and style, other papers focus on revising an entangled representation of an input. A few previous studies utilize a pre-trained classifier and edit entangled latent variable until it contains target style using the gradient-based optimization \cite{NIPS2019_9284,liu2019revision}. \newcite{he2020a} view each domain of data as a partially observable variable, and transfer sentence using amortized variational inference. \newcite{dai-etal-2019-style} use the transformer architecture and rewrite style in the entangled representation at the decoder. We consider this model as the strongest baseline model in terms of content preservation. 

In the domain of computer vision, it is a prevalent practice to exploit variants of normalization to transfer style \cite{DBLP:conf/iclr/DumoulinSK17, DBLP:journals/corr/UlyanovVL16}. \newcite{DBLP:conf/iclr/DumoulinSK17} proposed \textit{conditional instance normalization} (CIN) in which each style is assigned with separate instance normalization parameter, in other words, a model learns separate gain and bias parameters of instance normalization for each style. 

Our work differs in several ways. Style transfer in image views style transfer as changing the “texture” of an image. Therefore, \newcite{DBLP:conf/iclr/DumoulinSK17} place CIN module following every convolution layer, “painting” with style-specific parameters on the content representation. Therefore, the network passes on entangled representation of an image. Our work is different in that we disentangle content and style, thus we do not overwrite content with style-specific parameters. In addition, we apply CLN only once before passing it to decoder.


\section{Approach}
\subsection{Task Definition}
Let $\mathcal{D} = \{(\mathbf{x}_i, s_i)_{i=1}^N\}$ be a training corpus, where each $\mathbf{x}_i$ is a sentence, and $s_i$ is its style label.  Our experiments were carried on a sentiment analysis task, where there are two style labels, namely ``positive" and ``negative."

The task is to learn from $\mathcal{D}$ a model $\hat{\mathbf{x}}_{\hat{s}} = f_{\theta}(\mathbf{x}, \hat{s})$, with parameters $\theta$,  that takes
an input sentence $\mathbf{x}$ and a target style $\hat{s}$ as inputs, and outputs a new sentence $\hat{\mathbf{x}}_{\hat{s}}$  that is in the target style and retains the content information of $\mathbf{x}$.
\subsection{Model Overview}
\label{modelOverview}
We conduct this task in an unsupervised environment in which ground truth sentence $\mathbf{x}_{\hat{s}}$ is not provided. To achieve our goal, we employ a style classifier $s=C(\mathbf{x})$ that takes a sentence $\mathbf{x}$ as input and returns its style label. We pre-train such model on $\mathcal{D}$ and keep it frozen in the process of learning $f_{\theta}$.

Given the style classifier $C(\mathbf{x})$, our task becomes to learn a model $\hat{\mathbf{x}}_{\hat{s}} = f_{\theta}(\mathbf{x}, \hat{s})$ such that $C(\hat{\mathbf{x}}_{\hat{s}})= \hat{s}$. As such, the task is conceptually similar to adversarial attack: The input $\mathbf{x}$ is from the style class $s$, and we want to modify it so that it will be classified into the target style class $\hat{s}$.

The architecture of our model $f_{\theta}$ is shown in Figure \ref{modelStructure}, which will some times referred to as the generator network. It consists of an encoder, a stylizer and a decoder.  The encoder maps an input sequence $\mathbf{x}$ into a style-independent representation $\mathbf{z}_{\mathbf{x}}$. Particularly, the encoder has a style marker module that computes attention scores of input tokens, and it ``reverses" them to estimate the content information.
The reversed attention scores are applied to the token embedding $E(\mathbf{x})$ and the results $E'(\mathbf{x})$ are fed to bidirectional GRU to produce $\mathbf{z}_{\mathbf{x}}$.  

The stylizer takes a target style $\hat{s}$ and the content representation $\mathbf{z}_{\mathbf{x}}$ as inputs, and produces a content-related style representation $\mathbf{z}_{\hat{s}}$.
Finally, the decoder takes the content representation $\mathbf{z}_{\mathbf{x}}$ and style representation $\mathbf{z}_{\hat{s}}$ as inputs, and generates a new sequence $\hat{\mathbf{x}}_{\hat{s}}$. 

\subsection{Encoder}
\subsubsection{Style Marker Module}
Let $\mathbf{x}=[x_{1}, x_{2}, \mathellipsis , x_{T}]$ be a length $T$ sequence of input with a style $s$. The style marker module is pre-trained in order to calculate the amount of style information in each token in a given input. We use one layer of bidirectional GRU with attention \cite{yang-etal-2016-hierarchical}. Specifically,
\begin{equation}
	\mathbf{v}_{t} = tanh(\mathbf{W}_w\mathbf{h}_{t} + \mathbf{b}_w)
\end{equation}
\begin{equation}
	\mathbf{\alpha}_{t} = \frac{exp(\mathbf{v}_{t}^\mathsf{T}\mathbf{u}/\tau)}{\sum_{t=1}^T{exp(\mathbf{v}_{t}^\mathsf{T}\mathbf{u}}/\tau)}
\end{equation}
where $\mathbf{h}_t$ is the hidden representation from the bidirectional GRU at time step $t$. $\mathbf{u}$ is learnable parameters initialized with random weights, and $\tau$ denotes the temperature in softmax. When pre-training the style marker module, we construct a sentence representation by taking the weighted sum of the token representations with the weights being the attention scores, and feed the context vector to a fully-connected layer. 
\begin{equation}
	\mathbf{o} = \sum_{t=1}^T{\alpha_{t}\mathbf{h}_{t}}
\end{equation}
\begin{equation}
	\mathbf{p} = softmax(\mathbf{W}_{c}\mathbf{o}+\mathbf{b}_c)
\end{equation}
The cross-entropy loss is used to learn the parameters of the style marker module. The attention scores in the style marker indicate what tokens are important to style classification, and to what extent. Those scores will be ``reversed" in the next section to reveal the content information.  The fully-connected layer of the style marker module is no longer needed once the style marker module is trained. It is hence removed.


\subsubsection{Reverse Attention}
Using attention score from the pre-trained style marker module, we propose to implicitly remove the style information in each token. We negate the extent of style information in each token to estimate the extent of content information, namely reverse attention.
\begin{equation}
	\tilde{\alpha_t} = 1-\alpha_t, \quad \sum_{t=1}^{T} \alpha_{t} = 1
\end{equation}
where $\alpha_t$ is an attention value from style marker module, and $\tilde{\alpha}_t$ is the corresponding reverse attention score. We multiply the reverse attention scores to the embedding vectors of tokens.

\begin{equation}\label{eq:sentenceRe1}
\mathbf{\tilde{e}}_t =\tilde{\alpha_t}\mathbf{e}_t, \quad \mathbf{e}_t =E(\mathbf{x}_t)
\end{equation}
Intuitively, this can be viewed as implicitly removing the stylistic attribute of tokens, suppressing the norm of a token embedding respect to corresponding reverse attention score. The representations finally flow into a bidirectional GRU
\begin{equation}
\mathbf{z}_{\mathbf{x}} = bidirectionalGRU(\mathbf{\tilde{e}})
\end{equation}
to produce a content representation $\mathbf{z}_{\mathbf{x}}$, which is the last hidden state of the bidirectional GRU. By utilizing reverse attention, we map a sentence to style-independent content representation. 
\subsection{Stylizer}
The goal of the stylizer is to create a content-related style representation. We do this by applying conditional layer normalization on the content representation $\mathbf{z}_{\mathbf{x}}$ from encoder as input to this module. 

Layer normalization requires the number of gain and bias parameters to match the size of input representation. Therefore, mainly for the purpose of shrinking the size, we perform affine transformation on the content variable.
\begin{equation}\label{eq:conLN0}
\mathbf{\tilde{z}}_{\mathbf{x}} = \mathbf{W}_z\mathbf{z}_{\mathbf{x}}+\mathbf{b}_z
\end{equation}
The representation is then fed to conditional layer normalization so that the representation falls into target style distribution in style space.
Specifically,
\begin{equation}\label{eq:conLN1}
\mathbf{z}_{\hat{s}}= CLN(\mathbf{\tilde{z}}_{\mathbf{x}};\hat{s}) = \gamma^{\hat{s}}\odot N(\mathbf{\tilde{z}}_{\mathbf{x}})+\beta^{\hat{s}}
\end{equation}
\begin{equation}\label{eq:conLN2}
N(\mathbf{\tilde{z}}_{\mathbf{x}}) = \frac{\mathbf{\tilde{z}}_{\mathbf{x}}-\mu}{\sigma}
\end{equation}
where $\mu$ and $\sigma$ are mean and standard deviation of input vector respectively, and $\hat{s}$ is target style. Our model learns separate $\gamma^s$ (gain) and $\beta^s$ (bias) parameters for different styles. 

Normalization method is commonly used to change feature values in common scale, but known to implicitly keep the features. Therefore, we argue that the normalized content feature values retain content information of the content variable. By passing through conditional layer normalization module, the content latent vector is scaled and shifted with style-specific gain and bias parameter, falling into target style distribution. Thus, unlike previous attempts in text style transfer, the style representation is dynamic respect to the content, being content-dependent embedding.

In order to block backpropagation signal related to style flowing into $\mathbf{z}_{\mathbf{x}}$, we apply stop gradient on $\mathbf{z}_{\mathbf{x}}$ before feeding it to stylizer. 

\subsection{Decoder}
\label{decoder}
The decoder generates a sentence with the target style conditioned on content-related style representation and content representation. We construct our decoder using one single layer of GRU.
\begin{equation}\label{eq:Decoder}
\mathbf{\hat{x}}_{\hat{s}} \sim Dec_{\theta}(\mathbf{z}_{\mathbf{x}}, \mathbf{z}_{\hat{s}}) = p_{D}(\hat{\mathbf{x}}_{\hat{s}}|\mathbf{z}_{\mathbf{x}},\mathbf{z}_{\hat{s}})
\end{equation}
As briefly discussed in Section \ref{modelOverview}, the outputs from our generator are further passed on for different loss functions. However, sampling process or greedy decoding does not allow gradient to flow, because the methods are not differentiable. Therefore, we use soft sampling to keep the gradient flow. Specifically, when the gradient flow is required through the outputs, we take the product of probability distribution of each time step and the weight of embedding layer to project the outputs onto word embedding space. We empirically found that soft sampling is more suitable in our environment than gumbel-softmax \cite{Jang2017CategoricalRW}.
\subsection{Pre-trained Style Classifier}
\label{pretrainedClassifier}
Due to the lack of parallel corpus, we cannot train generator network with maximum likelihood estimation on style transfer ability. Therefore, this paper employs a pre-trained classifier $C(\mathbf{x})$ to train our generator on transferring style. Our classifier network has the same structure as style marker module with fully-connected layer appended, nonetheless, it is a separate model obtained from a different set of initial model parameters. We use the cross-entropy loss for training:

\begin{equation}\label{eq:TransferLoss}
\mathcal{L}_{pre} = - \mathbb{E}_{(\mathbf{x}, s) \sim \mathcal{D}}[\log p_C({{s}}|{\mathbf{x}}_{{s}})]
\end{equation}

We freeze the weights of this network after it has been fully trained.

\begin{figure}[t]
\centering
\includegraphics[width=0.9\columnwidth]{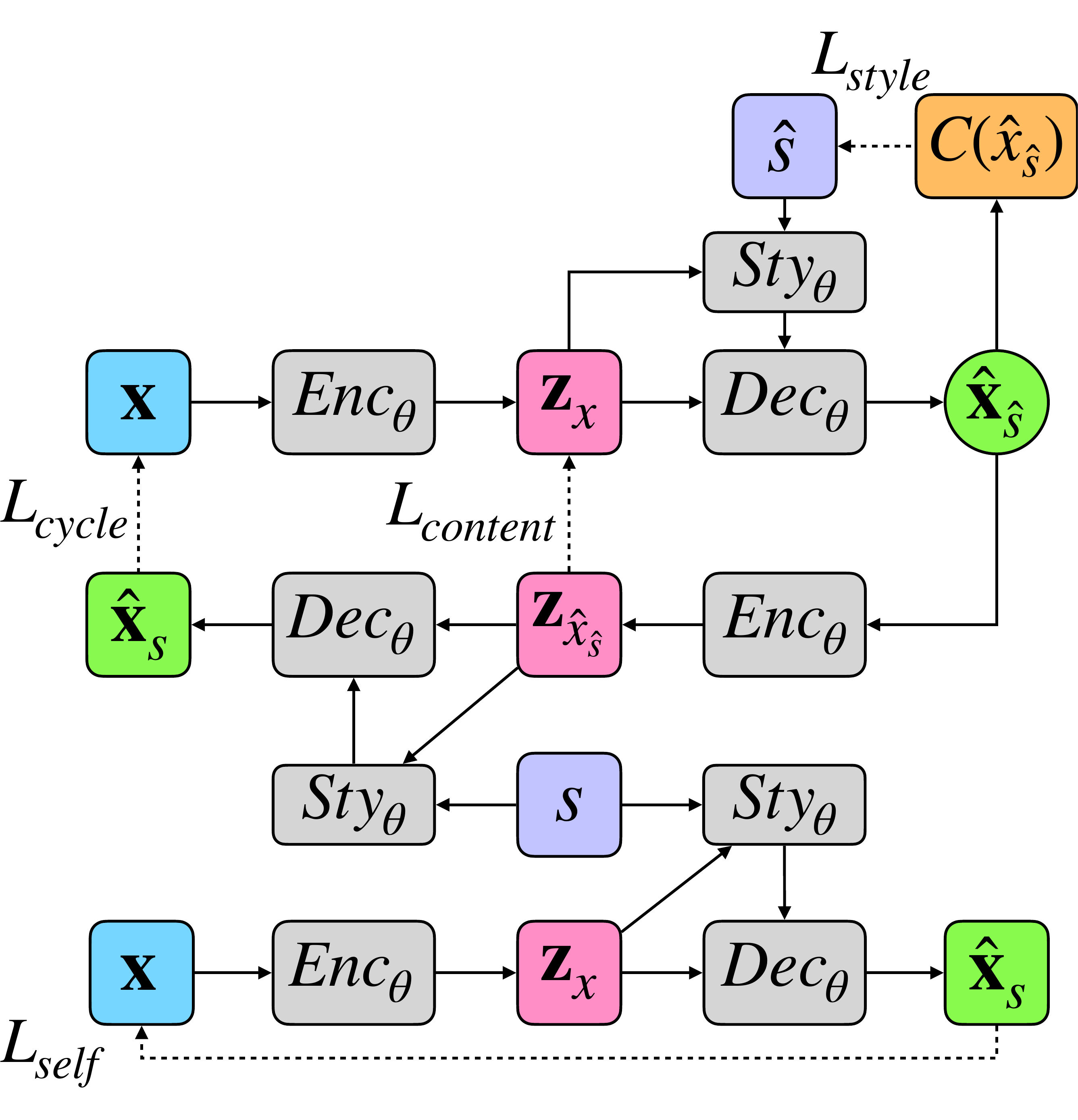}
\caption{\small Illustration of loss functions in training phase. $Enc_{\theta}, Sty_{\theta},$ and $Dec_{\theta}$ denote the encoder, the stylizer, and the decoder respectively. The circle figure denotes a generated sentence with soft sampling. As illustrated, $\mathcal{L}_{cycle}, \mathcal{L}_{style}$ and $\mathcal{L}_{content}$ require soft sampling to keep the gradient flow.}
\label{trainingLoss}
\end{figure}
\subsection{The Loss Function}
As shown in Figure \ref{trainingLoss}, our loss function consists of four parts: a self reconstruction loss $\mathcal{L}_{self}$, a cycle reconstruction loss $\mathcal{L}_{cycle}$, a content loss $\mathcal{L}_{content}$, and a style transfer loss $\mathcal{L}_{style}$.
\subsubsection{Self Reconstruction Loss}
Let $(\mathbf{x}, s) \in \mathcal{D}$ be a training example. If we ask our model to $f_{\theta}(\mathbf{x}, \hat{s})$ to ``transfer" the input into its original style, i.e., $\hat{s} = s$, we would expect it to reconstruct the input.
\begin{equation}
\mathcal{L}_{self} = - \mathbb{E}_{(\mathbf{x}, s) \sim \mathcal{D}}[\log p_D(\mathbf{x}|\mathbf{z}_{\mathbf{x}}, \mathbf{z}_{s})]
\label{self}
\end{equation}
where $\mathbf{z}_{\mathbf{x}}$ is the content representation of the input $\mathbf{x}$, $\mathbf{z}_{s}$ is the representation of the style $s$, and $p_D$ is the conditional distribution over sequences defined by the decoder. 
\subsubsection{Cycle Reconstruction Loss}
Suppose we first transfer a sequence  $\mathbf{x}$ into another style $\hat{s}$ to get $\hat{\mathbf{x}}_{\hat{s}}$ using soft sampling, and then transfer $\hat{\mathbf{x}}_{\hat{s}}$ back to the original style $s$.  We would expect to reconstruct the input $\mathbf{x}$.  Hence we have the following cycle construction loss:
\begin{equation}
\mathcal{L}_{cycle} = - \mathbb{E}_{(\mathbf{x}, s) \sim \mathcal{D}}[\log p_D(\mathbf{x}|\mathbf{z}_{\hat{\mathbf{x}}_{\hat{s}}}, \mathbf{z}_{s})]
\label{cycle}
\end{equation}
\noindent where $\mathbf{z}_{\hat{\mathbf{x}}_{\hat{s}}}$ is the content representation of the transferred sequence  $\hat{\mathbf{x}}_{\hat{s}}$.\footnote{Strictly speaking, the quantity is not well-defined because there is no description of how the target style $\hat{s}$ is picked. In our experiments, we use data with two styles. So, the target style just means the other style. To apply the method to problems with multiple styles, random sampling of different style should be added. This remark applies also to the two loss terms to be introduced below.} 
\subsubsection{Content Loss}
In the aforementioned cycle reconstruction process,  we obtain a content representation $\mathbf{z}_{\mathbf{x}}$ of the input $\mathbf{x}$ and a content representation  $\mathbf{z}_{\hat{\mathbf{x}}_{\hat{s}}}$ of the transferred sequence  $\hat{\mathbf{x}}_{\hat{s}}$.  As the two transfer steps presumably involve only style but not content, the two content representations should be similar. Hence we have the following content loss:
\begin{equation}\label{contentLossFunction}
\mathcal{L}_{content} = \mathbb{E}_{(\mathbf{x}, s) \sim \mathcal{D}} ||\mathbf{z}_{\mathbf{x}} - \mathbf{z}_{\hat{\mathbf{x}}_{\hat{s}}}||^2_2
\end{equation}
\subsubsection{Style Transfer Loss}
We would like the transferred sequence $\hat{\mathbf{x}}_{\hat{s}}$ to be of style ${\hat{s}}$. Hence we have the following style transfer loss:
\begin{equation}\label{styleTransfer}
\mathcal{L}_{style} =  - \mathbb{E}_{(\mathbf{x}, s) \sim \mathcal{D}}[\log p_C({\hat{s}}|\hat{\mathbf{x}}_{\hat{s}})]
\end{equation}
where $p_C$ is the conditional distribution over styles defined by the style classifier $C(\mathbf{x})$.  As mentioned in Section \ref{decoder}, $\hat{\mathbf{x}}_{\hat{s}}$ was generated with soft sampling. 
\subsubsection{Total Loss}
In summary, we balance the four loss functions to train our model. 
\begin{equation}
\mathcal{L} = \lambda_1\mathcal{L}_{self}+\lambda_2\mathcal{L}_{cycle}+\lambda_3\mathcal{L}_{content}+\lambda_4\mathcal{L}_{style}
\label{total}
\end{equation}
where $\lambda_i$ is balancing parameter.
\section{Experiment}
\subsection{Datasets}
Following prior work on text style transfer, we use two common datasets: Yelp and IMDB review.
\subsubsection{Yelp Review}
Our study uses Yelp review dataset \cite{li-etal-2018-delete} which contains 266K positive and 177K negative reviews. Test set contains a total of 1000 sentences, 500 positive and 500 negative, and human-annotated sentences are provided which are used in measuring content preservation.
\subsubsection{IMDB Movie Review}
Another dataset we test is IMDB movie review dataset \cite{dai-etal-2019-style}. This dataset is comprised of 17.9K positive and 18.8K negative reviews for training corpus, and 2K sentences are used for testing.
\subsection{Automatic Evaluation}
\subsubsection{Style Transfer Accuracy}
Style transfer accuracy (S-ACC) measures whether the generated sentences reveal target style property. We have mentioned a style classifier before: $C(\mathbf{x})$ which is used in the loss function. To evaluate transfer accuracy, we train another style classifier  $C_{eval}(\mathbf{x})$.  It has the identical architecture as before and trained on the same data, except from a different set of initial model parameters. We utilize such structure due to its superior performance compared to that of commonly used CNN-based classifier \cite{kim-2014-convolutional}. Our evaluation classifier achieves accuracy of 97.8\% on Yelp and 98.9\% on IMDB, which are higher than that of CNN-based.
\begin{table*}[t]
  \caption{\small Automatic evaluation result on Yelp dataset. Bold numbers indicate best performance. G-Score denotes geometric mean of self-BLEU and S-ACC, and BERT-P, BERT-R, and BERT-F1 are BERT score precision, recall and F1 respectively. All the baseline model outputs and codes were used from their official repositories if provided to the public.}\smallskip
  \centering
  \resizebox{\textwidth}{!}{
  \smallskip\begin{tabular}{l*{8}{c}}
  \toprule
  &\multicolumn{8}{c}{Yelp}\\
  & S-ACC& ref-BLEU& self-BLEU& PPL & G-score & BERT-P & BERT-R &BERT-F1\\
  \midrule
  Cross-Alignment \cite{NIPS2017_7259}\footnotemark & 74.2 	& 4.2& 13.2 	& 53.1& 32.0 & 87.8& 86.2 & 87.0\\
  ControlledGen \cite{pmlr-v70-hu17e}\footnotemark & 83.7 	& 16.1 	& 50.5 	& 146.3 & 65.0&90.6&89.0&89.8\\
  Style Transformer \cite{dai-etal-2019-style}\footnotemark & 87.3 	& 19.8 	& 55.2& 73.8&69.4&91.6&89.9&90.7\\
  Deep Latent \cite{he2020a}\footnotemark & 85.2 	& 15.1 	& 40.7& \textbf{36.7}& 58.9&89.8&88.6&89.2\\
  \hline
  RACoLN (Ours) & \textbf{91.3} & \textbf{20.0} & \textbf{59.4} & 60.1 & \textbf{73.6} & \textbf{91.8} & \textbf{90.3} & \textbf{91.0}\\
  \bottomrule
  \end{tabular}}
  \label{automaticEvaluation}
  \end{table*}

  \begin{table}[t]
  \caption{\small Automatic evaluation result on IMDB dataset. Bold numbers indicate best performance. As for IMDB Dataset, in the absence of human reference, BERT score and reference BLEU are not reported.}\smallskip
  \centering
  \resizebox{\columnwidth}{!}{
  \smallskip\begin{tabular}{l*{8}{c}}
  \toprule
  &\multicolumn{4}{c}{IMDB}\\
  & S-ACC& self-BLEU& PPL & G-score\\
  \midrule
  Cross-Alignment & 63.9&1.1&\textbf{29.9}&8.4\\
  ControlledGen& 81.2&63.8&119.7&71.2\\
  Style Transformer	& 74.0 	& 70.4 	& 71.2& 72.2\\
  Deep Latent 	& 59.3 	& 64.0 	& 41.1& 61.6\\
  \hline
  RACoLN (Ours) & \textbf{83.1} & \textbf{70.9} & 45.3 & \textbf{76.8}\\
  \bottomrule
  \end{tabular}}
  \label{automaticEvaluation-IMDB}
  \end{table}
  \begin{table}[t]
  \caption{\small Human evaluation result. Each score indicates the average score from the hired annotators. The inter-annotator agreement, Krippendorff's alpha, is 0.729.}\smallskip
  \centering
  \resizebox{\columnwidth}{!}{
  \smallskip\begin{tabular}{lcccccc}
  \toprule
  &\multicolumn{3}{c}{YELP}&\multicolumn{3}{c}{IMDB}\\
  \cmidrule(lr){2-4}\cmidrule(lr){5-7}
   & Style & Content & Fluency & Style & Content & Fluency\\
  \midrule
  Cross-Alignment & 2.6 & 2.4 & 3.3 & 2.2 & 2.1 & 2.3\\
  ControlledGen & 3.3 & 4.0 & 3.7 & 3.3 & 3.8 & 3.6\\
  Style Transformer& 3.7 & 4.3 & 4.0 & 3.3 & 4.0 & 3.8\\
  Deep Latent & 3.5 & 3.6 & \textbf{4.3} & 2.7 & 3.7 &\textbf{4.2}\\\hline
  RACoLN (Ours) & \textbf{4.0} & \textbf{4.5} & 4.2 &\textbf{3.6} & \textbf{4.1} & 4.1\\
  \toprule
  \end{tabular}
  }
  \label{humanEvaluation}
  \end{table}

\subsubsection{Content Preservation}
A well-transferred sentence must maintain its content. In this paper, content preservation was evaluated with two BLEU scores \cite{papineni-etal-2002-bleu}, one between generated sentence and input sentence (self-BLEU), and the other with human-generated sentence (ref-BLEU). With this metric, one can evaluate how a sentence maintains its content throughout inference.
\subsubsection{Fluency}
A natural language generation task aims to output a sentence, which is not only task-specific, but also fluent. This study measures perplexity (PPL) of generated sentences in order to measure fluency. Following \cite{dai-etal-2019-style}, we use 5-gram KenLM \cite{heafield-2011-kenlm} trained on the two training datasets. A lower PPL score indicates a transferred sentence is more fluent.

\subsubsection{BERT Score}
\citet{bert-score} proposed BERT score which computes contextual similarity of two sentences. Previous methods, such as BLEU score, compute n-gram matching score, while BERT score evaluates the contextual embedding of the tokens obtained from pre-trained BERT \cite{devlin-etal-2019-bert}. This evaluation metric has been shown to correlate with human judgement, thus our paper includes BERT score between model generated output and the human reference sentences. We report precision, recall, and F1 score. 

\subsection{Human Evaluation}
In addition to automatic evaluation, we validate the generated outputs with human evaluation. 
With each model, we randomly sample 150 outputs from each of the two datasets, total of 300 outputs per model.
Given the target style and the original sentence, the annotators are asked to evaluate the model generated sentence with a score range from 1 (Very Bad) to 5 (Very Good) on content preservation, style transfer accuracy, and fluency. We report the average scores from the 4 hired annotators in Table \ref{humanEvaluation}.
\subsection{Implementation Details}
In this paper, we set the embedding size to 128 dimension and hidden representation dimension of encoder to 500. The size of bias and gain parameters of conditional layer norm is 200, and the size of hidden representation for decoder is set to 700 to condition on both content and style representation. Adam optimizer \cite{DBLP:journals/corr/KingmaB14} was used to update parameter with learning rate set to 0.0005. For balancing parameters of total loss function, we set to 0.5 for $\lambda_1$ and $\lambda_2$, and 1 for the rest.
\footnotetext[2]{\url{https://github.com/shentianxiao/language-style-transfer}}
\footnotetext[3]{\url{https://github.com/asyml/texar/tree/master/examples/text_style_transfer}}
\footnotetext[4]{\url{https://github.com/fastnlp/style-transformer}}
\footnotetext[5]{\url{https://github.com/cindyxinyiwang/deep-latent-sequence-model}}
\subsection{Experimental Result \& Analysis}
We compare our model with the baseline models, and the automatic evaluation result is presented in Table \ref{automaticEvaluation}. Our model outperforms the baseline models in terms of content preservation on both of the datasets. Especially, on Yelp dataset, our model achieves 59.4 self-BLEU score, surpassing the previous state-of-the-art model by more than 4 points. Furthermore, our model also achieves the state-of-the-art result in content preservation on IMDB dataset, which is comprised of longer sequences than those of Yelp.

In terms of style transfer accuracy and fluency, our model is highly competitive. Our model achieves the highest score in style transfer accuracy on both of the datasets (91.3 on Yelp and 83.1 on IMDB). Additionally, our model shows the ability to produce fluent sentences as shown in the perplexity score. In terms of the BERT scores, the proposed model performs the best, having the highest contextual similarity with the human reference among the style transfer models. 

With the automatic evaluation result, we see a trend of trade-off. Most of the baseline models are good at particular metric, but show room for improvement on other metrics. For example, Deep Latent and Cross-Alignment constantly perform well in terms of perplexity, but their ability to transfer style and preserving content needs improvement. Style Transformer achieves comparable performance across all evaluation metrics, but our model outperforms the model on every metric on both of the datasets. Therefore, the result shows that our model is well-balanced but also strong in every aspect in text style transfer task.

As for the human evaluation, we observe that the result mainly conform with the automatic evaluation. Our model received the highest score on the style and content evaluation metric on both of the datasets by a large margin compared to the other baselines. Moreover, the fluency score is comparable with that of Deep Latent model, showing its competency in creating a fluent output. Both automatic and human evaluation depict the strength of the proposed model not only in preserving content, but also on other metrics. 

\begin{table}[t]
\caption{\small Sample outputs generated by the baseline models and our approach on Yelp and IMDB dataset. Bold words indicate successful transfer in style without grammatical error.}\smallskip
\centering
\resizebox{\columnwidth}{!}{
\smallskip\begin{tabular}{ll}
\toprule
 				\multicolumn{2}{c}{YELP}\\
\midrule
Original Input & Everyone is always super friendly and helpful .\\\hdashline
\multirow{2}{*}{Cross-Alignment} & Everyone is always super friendly \\&and helpful and inattentive . \\\hdashline
ControlledGen &Tonight selection of meats and cheeses . \\\hdashline
Deep Latent &Now i 'm not sure how to be . \\\hdashline
Style Transformer 		& Which is \textbf{n't} super friendly . \\\hdashline
RACoLN (Ours) & Everyone is always super \textbf{rude} and \textbf{unprofessional} . \\
\midrule
Original Input & I love this place , the service is always great !\\\hdashline
Cross-Alignment & I know this place , the food is just a horrible ! \\\hdashline
ControlledGen &I \textbf{avoid} this place , the service is nasty depressing vomit \\\hdashline
Deep Latent &I do n't know why the service is always great ! \\\hdashline
Style Transformer 		& I \textbf{do n't} recommend this place , the service is n't ! \\\hdashline
RACoLN (Ours) & I \textbf{avoid} this place , the service is always \textbf{horrible} ! \\\\
\toprule
 					\multicolumn{2}{c}{IMDB}\\
\midrule
\multirow{2}{*}{Original Input} & I actually disliked the leading characters so much \\&that their antics were never funny but pathetic .\\\hdashline
\multirow{2}{*}{Cross-Alignment} & I have never get a good movie , i have never have\\&seen in this movie . \\\hdashline
\multirow{2}{*}{ControlledGen} &I actually anticipated the leading characters so much\\&that their antics were never funny but timeless .\\\hdashline
\multirow{2}{*}{Deep Latent} &I actually disliked the leading characters so much\\&that their antics were never funny but \textbf{incredible} . \\\hdashline
\multirow{2}{*}{Style Transformer} 		& I actually disliked the leading characters so much \\&that their antics were never funny but vhs .  \\\hdashline
\multirow{2}{*}{RACoLN (Ours)}  & I actually \textbf{liked} the leading characters so much \\&that their antics were \textbf{never corny but appropriate} . \\
\midrule
Original Input & The plot is clumsy and has holes in it .\\\hdashline
Cross-Alignment & The worst film is one of the worst movies i 've ever seen . \\\hdashline
ControlledGen &The plot is \textbf{top-notch} and has one-liners in it . \\\hdashline
Deep Latent &The plot is tight and has found it in a very well done . \\\hdashline
Style Transformer  		& The plot is joys and has flynn in it .\\\hdashline
RACoLN (Ours) & The plot is \textbf{incredible} and has \textbf{twists} in it . \\
\end{tabular}
}
\label{sample}
\end{table}
\begin{figure}[ht]
\centering
\begin{minipage}[b]{0.45\linewidth}
\includegraphics[width=\textwidth]{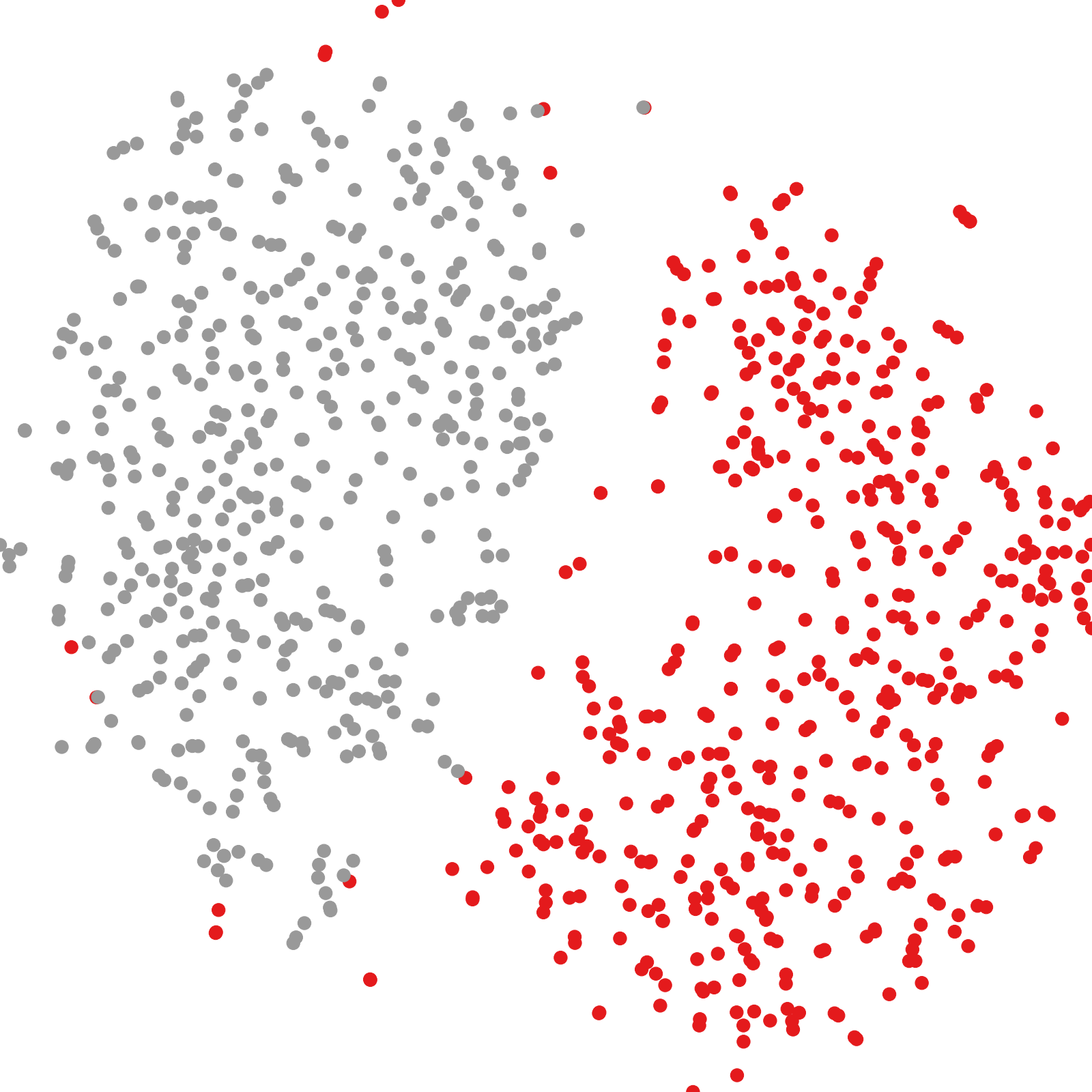}
\caption*{Style Space}
\label{fig:minipage1}
\end{minipage}
\quad
\begin{minipage}[b]{0.45\linewidth}
\includegraphics[width=\textwidth]{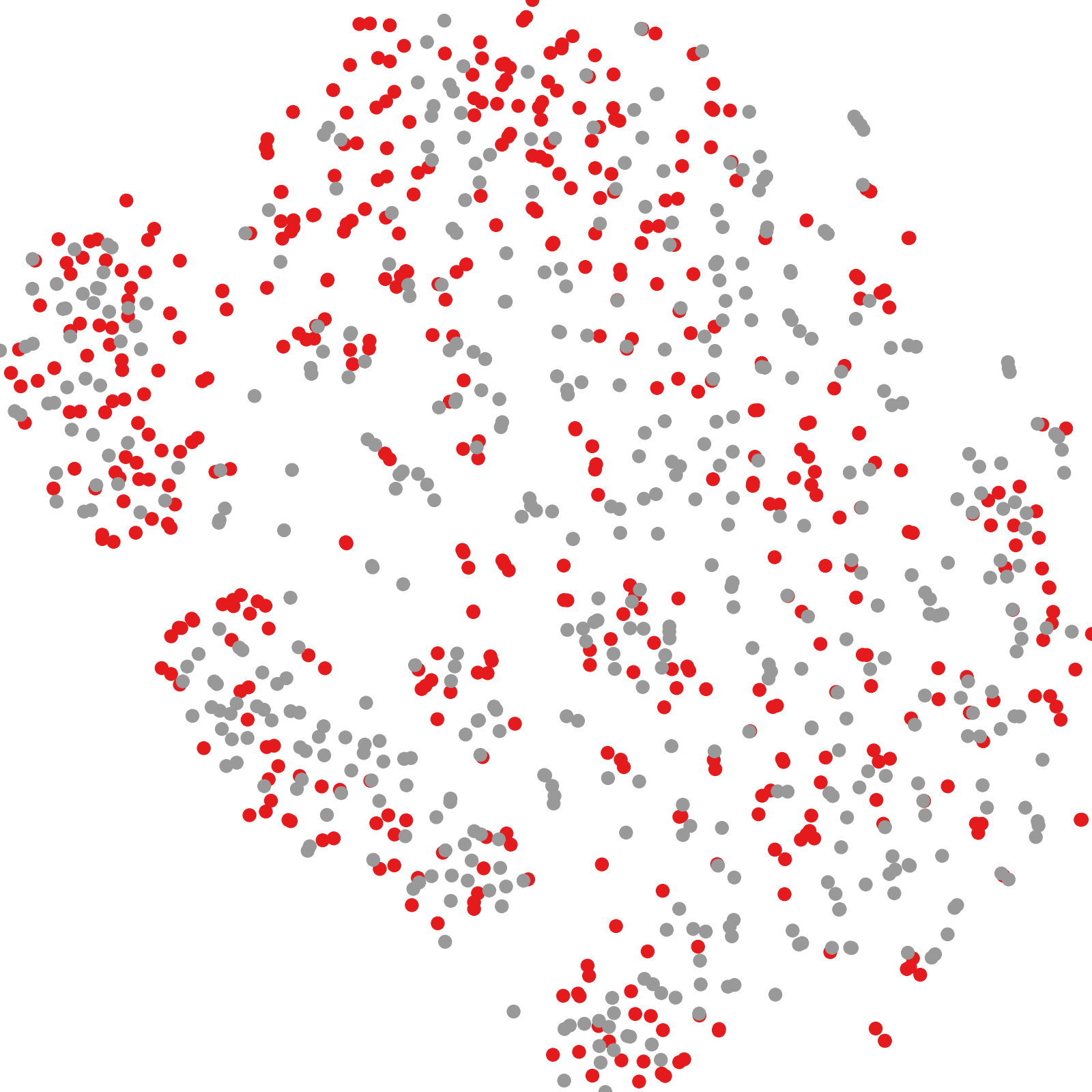}
\caption*{Content Space}
\label{fig:minipage2}
\end{minipage}
\caption{\small Visualization of Yelp test dataset on content and style space using t-SNE. Gray dots denote sentences with negative style transferred to positive sentiment, while red dots are sentences with positive style transferred to negative sentiment.}
\label{fig:spaceFigure}
\end{figure}
\subsubsection{Style and Content Space}
We visualize the test dataset of Yelp projected on content and style space using t-SNE in Figure \ref{fig:spaceFigure}. It is clearly observed that the content representations ($\mathbf{z}_{\mathbf{x}}$) are spread across content space, showing that the representations are independent of style. After the content representations go through the stylizer module, there is a clear distinction between different styles representations ($\mathbf{z}_{\hat{s}}$) in style space. This is in sharp contrast to the corresponding distributions of the style-independent content representations shown on the right of the figure. The figure clearly depicts how style-specific parameters in the stylizer module shape the content representations to fall in the target style distribution. This figure illustrates how our model successfully removes style at the encoder, and constructs content-related style at the stylizer module.  
\begin{table}[t]
\caption{\small Ablation study on the proposed model. (-) indicates removing the corresponding component from the proposed model.}\smallskip
\centering
\resizebox{\columnwidth}{!}{%
\smallskip\begin{tabular}{l|c c c c}
\toprule
& S-ACC& ref-BLEU & self-BLEU & PPL\\
\midrule
Input Copy & 2.2 & 22.7 & 100.0 & 41.2\\
\midrule
Proposed Model & 91.3 & 20.0 	& 59.4 & 60.1\\
\midrule
(-) Reverse Attention & 84.0 & 16.6 & 47.2 & 60.5 \\
(-) Stylizer & 91.8 & 19.1 & 53.0 & 59.0 \\
\hdashline
(-) $\mathcal{L}_{content}$ & 87.2 & 19.5 & 54.8 & 62\\
\bottomrule
\end{tabular}}
\label{ablation}
\end{table}
\subsubsection{Ablation Study}
In order to validate the proposed modules, we conduct ablation study on Yelp dataset which is presented in Table \ref{ablation}. We observe a significant drop across all aspects without the reverse attention module. In other case, where we remove the stylizer module and use style embedding as in the previous papers, the model loses the ability to retain content, drop of around 6 score on self-BLEU. We find that the two core components are interdependent in successfully transferring style in text. Lastly, as for the loss functions, incorporating $\mathcal{L}_{content}$ brings a meaningful increase in content preservation.\footnote{Other loss functions were not included, since the loss functions have been extensively tested and explored in previous papers \cite{prabhumoye-etal-2018-style,dai-etal-2019-style}.}
\section{Conclusion}
In this paper, we introduce a way to implicitly remove style at the token level using reverse attention, and fuse content information to style representation using conditional layer normalization. With the two core components, our model is able to enhance content preservation while keeping the outputs fluent with target style. Both automatic and human evaluation shows that our model has the best ability in preserving content and is strong in other metrics as well. In the future, we plan to study problems with more than two styles and apply multiple attribute style transfer, where the target style is comprised of multiple styles.
\section*{Acknowledgement}
Research on this paper was supported by Hong Kong Research Grants Council under grant 16204920 and Tencent AI Lab Rhino-Bird Focused Research Program (No. GF202035).
\section*{Ethical Considerations}
A text style transfer model is a conditional generative model, in which the condition is the target style. 
This makes a wide range of applications possible, since a style can be defined as any common feature in a corpus, such as formality, tense, sentiment, etc.

However, at the same time, due to its inherent functionality, a text style transfer model can pose potential harm when used with a malicious intention. 
It can lead to a situation where one deliberately distorts a sentence for his or her own benefit.
To give an example in a political context, political stance can be viewed a style in political slant dataset \cite{voigt-etal-2018-rtgender} as in \cite{prabhumoye-etal-2018-style}.
If one intentionally changes the style (political stance) of a person with the proposed model structure, the generated output can be exploited to create fake news or misinformation.
One possible remedy for such potentially problematic situation is to employ fact checking system as a safety measure \cite{nadeem-etal-2019-fakta}.
We are fully aware that fact checking is not the fundamental solution to the potential harm that text style transfer models possess. Nevertheless, one can filter out misleading information using the system in certain domains (i.e., politics), lowering the level of the danger that can be otherwise posed by style transfer.
In conclusion, such problem is shared among conditional generative models in general, and future studies on how to mitigate this problem are in crucial need.


Our work validates the proposed model and the baseline models on human evaluation, in which manual work was involved. 
Thus, we disclose the compensation level given to the hired annotators. 
The average lengths of the two corpora tested are 10.3 words for Yelp and 15.5 words for IMDB. 
In addition, the annotation was performed on sentence-level, in which the annotators were asked to score a model generated sentence.
Considering the length and the difficulty, the expected annotations per hour was 100 sentences. 
The hourly pay was set to 100 Hong Kong dollars (HK\$), which is higher than Hong Kong's statutory minimum wage. 
The annotators evaluated 1,500 sentences in total (750 sentences per dataset), thus each annotator was compensated with the total amount of HK\$1,500.
\bibliographystyle{acl_natbib}
\bibliography{anthology,acl2021}
\end{document}